%% file: main.tex
\documentclass[11pt]{article}
\usepackage[preprint]{acl}
\usepackage{times}
\usepackage{latexsym}
\usepackage[T1]{fontenc}
\usepackage[utf8]{inputenc}
\usepackage[protrusion=true,expansion=false]{microtype}
\IfFileExists{inconsolata.sty}{\usepackage{inconsolata}}{}
\usepackage{graphicx,amsmath,amssymb,amsthm}
\usepackage{float,booktabs,multirow,tabularx,longtable,array,colortbl}
\usepackage{placeins,tikz,xcolor,needspace,multicol}
\makeatletter
\providecommand{\insert@pcolumn}{\insert@column}
\makeatother
\usepackage[skins,breakable]{tcolorbox}
\usetikzlibrary{arrows.meta,calc,positioning,fit}
\definecolor{ResultTeal}{RGB}{30,101,107}
\definecolor{ResultTealLight}{RGB}{241,248,247}
\definecolor{MethodOchre}{RGB}{123,83,45}
\definecolor{MethodOchreLight}{RGB}{251,246,239}
\usepackage{algpseudocode}
\floatstyle{ruled}
\newfloat{algorithm}{tbp}{loa}
\floatstyle{plain}
\floatname{algorithm}{Algorithm}
\newcolumntype{Y}{>{\raggedright\arraybackslash}X}

\newcommand{\appsec}[2]{\section{#1}\label{#2}\setcounter{equation}{0}}
\newtcolorbox{evidencebox}[1]{colback=ResultTealLight,colframe=ResultTeal!35,boxrule=.5pt,arc=1pt,left=5pt,right=5pt,top=4pt,bottom=4pt,fonttitle=\bfseries,title={#1},coltitle=black,colbacktitle=ResultTeal!12,breakable}
\title{Trace-to-Architecture:\\ Capability Bottleneck Diagnosis for Harness Revision}
\author{
  \textbf{Wenfeng Wang}\textsuperscript{1}\thanks{Equal contribution.} \quad
  \textbf{Jianrui Shi}\textsuperscript{2}\footnotemark[1]\thanks{Corresponding author.} \quad
  \textbf{Mingang Guo}\textsuperscript{1} \quad
  \textbf{Jiayuan Ma}\textsuperscript{2} \\
  \textbf{Yuqi Lu}\textsuperscript{2} \quad
  \textbf{Jiajun Sun}\textsuperscript{2} \quad
  \textbf{Guojiang Zhao}\textsuperscript{3} \quad
  \textbf{Jing Li}\textsuperscript{2} \\[0.6em]
  \normalfont\normalsize\textsuperscript{1}University of Pennsylvania, Philadelphia, USA \\
  \normalfont\normalsize\textsuperscript{2}Harbin Institute of Technology, Shenzhen, China \\
  \normalfont\normalsize\textsuperscript{3}DP Technology, Beijing, China \\[0.5em]
  {\normalfont\footnotesize
  \begin{tabular}{@{}ll@{}}
    \texttt{fw93810@seas.upenn.edu} &
    \texttt{jianrui.shi@stu.hit.edu.cn}
  \end{tabular}}
}
\hypersetup{
  pdftitle={Trace-to-Architecture: Capability Bottleneck Diagnosis for Harness Revision},
  pdfauthor={Wenfeng Wang, Jianrui Shi, Mingang Guo, Jiayuan Ma, Yuqi Lu, Jiajun Sun, Guojiang Zhao, Jing Li}
}
\begin{document}
\maketitle
\input{base/0_abstract}
\input{base/1_introduction}
\input{base/2_background}
\input{base/3_method}
\input{base/4_experiment}
\input{base/5_conclusion}
\input{base/6_limitations}
\bibliography{references}
\input{base/appendix}
\end{document}

%% file: base/0_abstract.tex
\begin{abstract}
Improving an LLM agent requires knowing what to change when it fails. Similar failures can require different fixes, and resolving one problem can expose another. We introduce Trace-to-Architecture (T2A), a framework that uses execution trajectories to determine what to change in an agent's harness and when those changes apply. A development agent diagnoses failures using execution evidence from successful and failed tasks, revises the corresponding harness components, and then checks whether target tasks improve, selected successful tasks remain successful, and the edits are actually triggered. New execution evidence guides subsequent diagnosis and revision. Across ALFWorld, ScienceWorld, and AppWorld, T2A achieves the highest aggregate score for all three evaluated task agents, exceeding their respective strongest baselines by 6.2--30.3 points. We also integrate T2A's diagnosis and validation procedures into Meta-Harness, improving the resulting harnesses' mean ScienceWorld score by 21.6 points from the same starting harness.
\end{abstract}

%% file: base/1_introduction.tex
\section{Introduction}
\label{sec:intro}
\label{sec:introduction}

Recursive self-improvement (RSI) aims to let agents turn accumulated experience into lasting improvements in their capabilities and in how they pursue further improvement \citep{zelikman2024stop,zhang2026dgm,duan2026lastai,zheng2026dreamrsi}. For agents built around a large language model (LLM), one practical route is to revise the agent harness: components such as prompts, context management, stored memory, and tool interfaces. These components govern how the model interacts with its environment. Revising the harness can improve an agent's task performance without updating model weights. MetaRSI incorporates such revisions alongside changes to data and model parameters \citep{tan2026metarsi}, while other systems automate harness revision using execution feedback \citep{lee2026metaharness,zhang2026selfharness,xia2026rrsi}. SoL-Pi further demonstrates how improvements to execution and context handling can reduce inference cost \citep{liu2026solpi}.

Optimizing a harness requires choosing among these options based on the current execution problems. Task outcomes alone provide limited guidance for this decision. Similar failures may arise from causes such as missing knowledge, failed delivery of memory, or inadequate execution rules, each requiring a different revision. Fixing one problem can also expose another or disrupt previously successful behavior. Records of model inputs and outputs, actions, and environment feedback help distinguish these causes \citep{barke2026agentrx,chen2026traceelephant}. Effective harness revision therefore requires checking diagnoses across related tasks, validating the effects of edits, and using new execution evidence to identify the next problem to address.

To this end, we introduce \textbf{Trace-to-Architecture (T2A)}, a framework for identifying what to change in a harness and checking whether those changes address the diagnosed failures. Within T2A, a task agent runs tasks under the current harness. A development agent compares related successful and failed execution trajectories to identify failure causes and determine which harness components to change and when those changes apply. Before evaluation, the agent records acceptance conditions covering target-task improvement, preservation of selected successful tasks, and execution evidence that the edits were triggered. All three are required for acceptance. Changes that pass validation are retained, and new execution trajectories help identify the next problems to address.

We evaluate T2A with three task agents in ALFWorld for household tasks, ScienceWorld for scientific experiments, and AppWorld for cross-application tasks \citep{shridhar2021alfworld,wang2022scienceworld,trivedi2024appworld}. T2A achieves the highest aggregate score across these environments for each task agent among the evaluated methods. With the starting harness and Meta-Harness's outer evaluation and selection procedure fixed, integrating T2A's diagnosis and validation requirements raises the resulting harnesses' mean ScienceWorld score by 21.6 points.

%% file: base/2_background.tex
\section{Related Work}
\label{sec:related}

\paragraph{Agent and harness improvement.}
Self-improving agents revise programs, experience, or optimization strategies \citep{zelikman2024stop,zhang2026dgm,tan2026metarsi}. With model weights fixed, GEPA adapts prompts using execution feedback \citep{agrawal2026gepa}, while Reflexion, CLIN, and WorldEvolver update experience or memory \citep{shinn2023reflexion,majumder2024clin,zhang2026worldevolver}. ADAS and AFlow search over agent programs and workflows \citep{hu2025adas,zhang2025aflow}. Meta-Harness extends this search to harness code, using historical implementations, scores, and trajectories to propose revisions \citep{lee2026metaharness}.

\paragraph{Diagnosis and evaluation.}
AgentRx, TraceElephant, and AgentDebugX study how execution records can support failure localization, attribution, and recovery \citep{barke2026agentrx,chen2026traceelephant,zhu2026agentdebugx}. Harness evaluation also distinguishes producing updates from benefiting from them \citep{lin2026harnessbenefit} and tests generalization under matched development budgets \citep{wang2026harnessevaluation}. Accordingly, we evaluate frozen-system task outcomes, changes in target errors, and development protocols under controlled conditions.

\paragraph{Harness revision.}
Prior work uses failure diagnosis to guide system modifications. Self-Harness summarizes recurring causes of failure and restricts changes to relevant components \citep{zhang2026selfharness}. Agentic Harness Engineering and HarnessX record the reasons and expected effects of changes and check those changes against execution evidence and task outcomes \citep{lin2026ahe,chen2026harnessx}. HarnessFix locates the steps and system components responsible for errors and checks whether repairs reduce the targeted errors \citep{chen2026harnessfix}. T2A likewise revises harnesses based on execution evidence, emphasizing the diagnosis of capability bottlenecks across tasks and the conditions under which they arise to guide revisions and formulate validation conditions. Appendix~\ref{app:revision-comparison} compares the workflows in detail.

%% file: base/3_method.tex
\section{Method}
\label{sec:method}
\subsection{Overview and Roles}
T2A alternates between \emph{data collection} (Section~\ref{sec:collection}) and \emph{harness revision} (Section~\ref{sec:revision}); Figure~\ref{fig:method} shows the process, and Algorithm~\ref{alg:t2a} in Appendix~\ref{app:t2a-workflow} summarizes the complete loop. In each round, the task agent runs tasks under the current harness and produces execution trajectories. The development agent then analyzes these trajectories and revises the harness. Development is bounded by the stopping conditions and resource limits specified for each experiment. Appendices~\ref{app:baseline-settings} and~\ref{app:development} describe the starting-harness requirements and task splits.

\begin{figure*}[t]
\centering
\includegraphics[width=\textwidth]{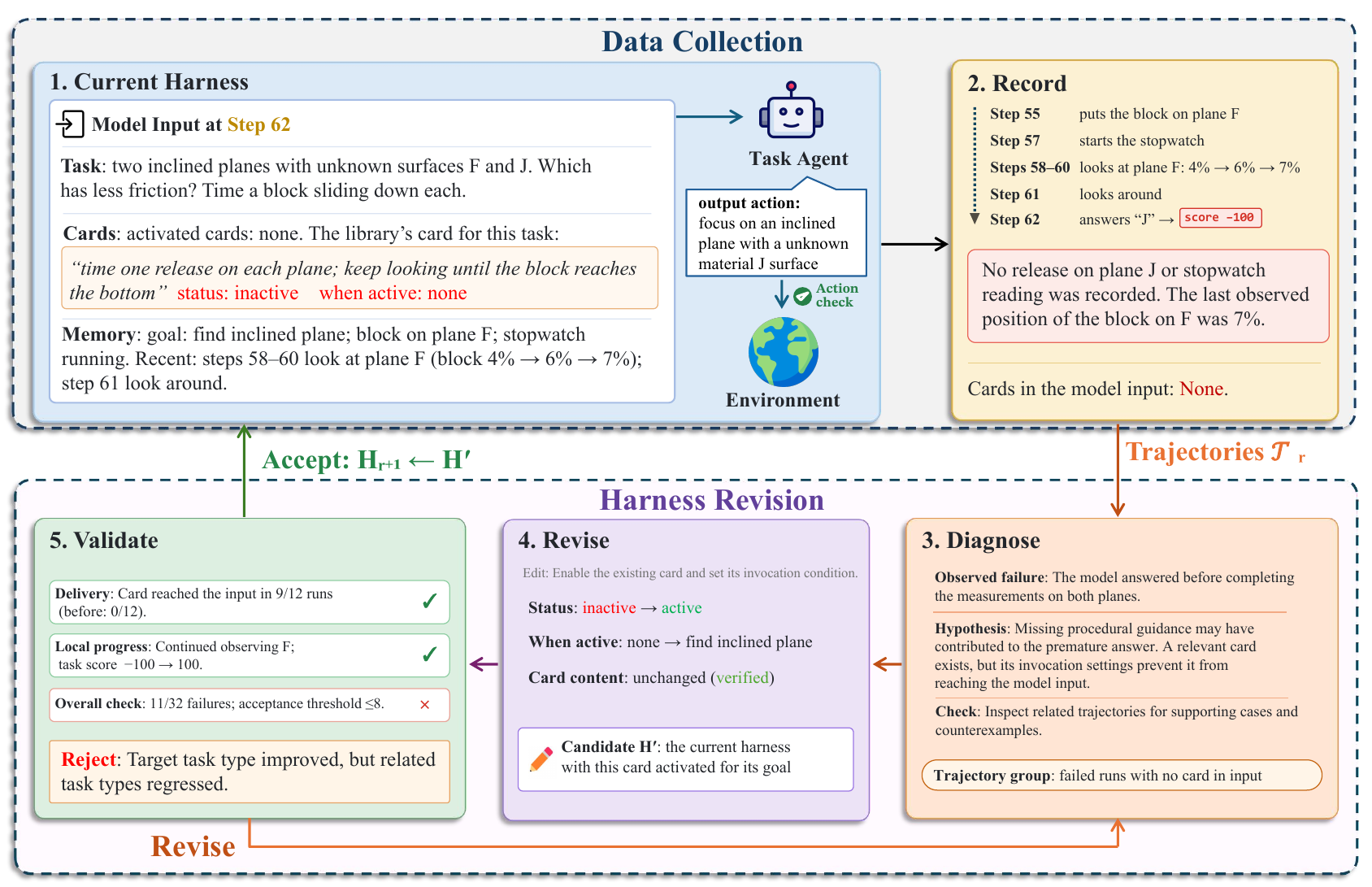}
\caption{T2A alternates between trajectory collection (top) and diagnosis, revision, and validation (bottom). In the example, execution evidence identifies a skill card that is absent from the model input. Adjusting its invocation conditions improves the target task, but regressions on related tasks lead to rejection and further revision.}
\label{fig:method}
\end{figure*}

\subsection{Data Collection: Execution and Trajectory Collection}
\label{sec:collection}
In round $r$, tasks run under a fixed harness
\begin{equation}
H_r=(D_r,\mathcal C_r,R_r).
\label{eq:configuration}
\end{equation}
The driver $D_r$ executes components, constructs model inputs from the current task state, and processes proposed actions before execution. Its editable scope includes input construction and action processing, but excludes the scorer and externally enforced resource limits. $\mathcal C_r$ contains editable components and their invocation conditions, such as skill cards, within-episode memory, and action checks. Skill cards are texts that store task knowledge or operational experience. When their invocation conditions are met, relevant cards are added to the model input. $R_r$ specifies what information is recorded during execution and when it is recorded.

For task $d$, the interaction records $e_{d,t}$ collected according to $R_r$, together with the terminal score $s_d$, form a trajectory:
\begin{equation}
\begin{aligned}
\tau_d&=((e_{d,1},\ldots,e_{d,n_d}),s_d),\\
\mathcal T_r&=\{\tau_d:d\in\mathcal Q_r\}.
\end{aligned}
\label{eq:traces}
\end{equation}
$\tau_d$ denotes the execution trajectory for task $d$, and $\mathcal T_r$ is the set of trajectories collected in round $r$. $\mathcal Q_r$ is the development task set used in that round, and $n_d$ is the number of interactions for task $d$.

\subsection{Harness Revision}
\label{sec:revision}

\paragraph{Diagnosis.}
In each round, T2A selects all tasks that did not receive full scores, giving $\mathcal F_r=\{\tau_d\in\mathcal T_r:s_d<s_d^{\max}\}$, where $s_d^{\max}$ is the full score for task $d$. Scripts group these trajectories into $\mathcal G_r$ using observable execution features, retaining task identifiers and execution steps for context retrieval. The development agent examines representative trajectories from each group, using model inputs, decisions, action processing, and environment feedback to identify possible causes of failure and the situations in which they apply. These form the candidate diagnosis set $\mathcal Z_r^{\mathrm{cand}}$. Each diagnosis record takes the form
\begin{equation}
z=(h_z,c_z,\mathcal E_z).
\label{eq:diagnosis-record}
\end{equation}
Here $h_z$ is a hypothesis about the cause of failure, describing how the observed problem relates to the information available at that point or to harness operations; $c_z$ specifies its applicability conditions; and $\mathcal E_z$ contains the concrete execution records used to check the hypothesis across related successful and failed tasks.

The agent searches the execution trajectory set $\mathcal T_r$ for related cases and checks each hypothesis in its execution context. The agent revises or withdraws explanations according to the evidence; problems whose causes cannot be determined from the available records remain unresolved. Diagnoses supported by execution records are consolidated for the same problem and kept separate for distinct problems, forming $\mathcal Z_r$. Every task that lost points is assigned to a diagnosis cluster or marked as unresolved; unresolved problems retain open questions and the additional records needed. The resulting records link diagnoses to relevant tasks and steps to guide revision. For example, in Figure~\ref{fig:method}, the required guidance is stored in a skill card, but the card is disabled for the current invocation and absent from the model input, identifying invocation settings as the revision target.

\paragraph{Revision.}
The development agent translates supported diagnoses $\mathcal Z_r$ into a harness edit $\delta$, linking the edit to the affected tasks and execution steps and describing its expected effects. Applying $\delta$ to the current harness $H_r$ produces a candidate harness $H'$ by modifying the driver, components, or logging rules. Component edits may modify existing components or introduce new ones. The recorded links between diagnoses, edits, and expected effects guide subsequent validation.

Revision targets depend on where execution breaks down. Possible edits include:
\begin{itemize}
\setlength{\itemsep}{0pt}
\setlength{\parskip}{0pt}
\item \textbf{Skill-card content:} Revise the card when knowledge is missing or its procedure is unsuitable.
\item \textbf{Guidance delivery:} Adjust retrieval or invocation conditions when applicable guidance is absent from the model input.
\item \textbf{Decision rules:} Clarify operating rules when decision errors recur.
\item \textbf{Action check:} Add checks based on observable prerequisites for invalid actions.
\item \textbf{Memory and execution control:} Adjust memory updates or execution control when observations are outdated or known states are repeatedly visited.
\end{itemize}
When the available records leave the cause unresolved, the agent revises $R_r$ to collect additional execution records or runs a targeted comparison before choosing a corrective change.
Appendix~\ref{app:lifespan-revision} illustrates how diagnosis and validation shift a revision from skill-card content to execution control.

\paragraph{Validation.}
Before running validation, the development agent records acceptance conditions for comparing the candidate harness $H'$ with the current harness $H_r$. These specify the target tasks and their expected results, protected tasks that currently succeed and may be affected by the edit, and how to verify that the edit was actually triggered. Checks of specific target behaviors are included when specified by the validation plan.

Using execution records from both harnesses, the development agent judges the candidate against these conditions:
\begin{itemize}
\setlength{\itemsep}{0pt}
\setlength{\parskip}{0pt}
\item \textbf{Target-task improvement:} Whether the target tasks improve as specified in the acceptance conditions.
\item \textbf{Regression protection:} Whether the protected tasks retain their successful outcomes.
\item \textbf{Edit triggering:} Whether execution records confirm that the edit was actually triggered, through component invocation, model inputs, or the relevant code path.
\end{itemize}
A candidate that fails validation may be retained for further revision or validation without being accepted. Acceptance requires satisfying all conditions of the applicable plan, registered before that validation run. Until acceptance, $H_r$ remains the accepted harness. The development agent writes the acceptance conditions and judges whether they are met from execution records, without researcher review; the program does not independently adjudicate semantic acceptance. The agent saves the validation plan, execution records, and verdict whether or not the candidate is accepted. After acceptance, trajectories produced by the new harness are used to begin the next round of diagnosis.

%% file: base/4_experiment.tex
\section{Experiments}
\label{sec:experiments}
\input{base/main_table}
We evaluate T2A with Qwen3.5-27B, Qwen3.5-9B \citep{qwen2026modelcards}, and DeepSeek V4.1 Flash \citep{deepseek2026v41} on three interactive environments: ALFWorld \citep{shridhar2021alfworld} for household tasks, ScienceWorld \citep{wang2022scienceworld} for scientific experiments, and AppWorld \citep{trivedi2024appworld} for cross-application tasks.

\subsection{Experimental Setup}
\label{sec:setup}
\paragraph{Environments and metrics.}
ALFWorld \citep{shridhar2021alfworld} requires agents to navigate rooms, find objects, and manipulate them to satisfy household goals, such as heating an object and placing it in a specified receptacle. ScienceWorld \citep{wang2022scienceworld} requires agents to locate materials, operate equipment, and use experimental observations to complete scientific tasks and select answers. AppWorld \citep{trivedi2024appworld} requires agents to write Python code to query and modify data across applications according to user requests. ALFWorld and AppWorld report task completion rates; ScienceWorld reports mean terminal score and fatal-failure rate.

\paragraph{Baselines.}
Minimal prompting supplies the task and interaction conventions. All methods in Table~\ref{tab:main} use the minimal-prompt harness as their starting point. ReAct \citep{yao2023react} uses a fixed reasoning--action--observation loop. WorldEvolver \citep{zhang2026worldevolver}, SkillNet \citep{liang2026skillnet}, GEPA \citep{agrawal2026gepa}, and Meta-Harness \citep{lee2026metaharness} represent memory, skill, prompt, and harness approaches. WorldEvolver is adapted to build memory on development tasks, freeze it, and execute each test task once. Baseline adaptations are in Appendix~\ref{app:baseline-settings}.

\paragraph{Implementation details.}
We use Qwen3.5-27B and Qwen3.5-9B \citep{qwen2026modelcards}, and DeepSeek V4.1 Flash \citep{deepseek2026v41} in all three environments to cover different levels of baseline task performance and examine how the initial distribution of capability bottlenecks affects subsequent revision outcomes. The development agent uses Claude Opus 5 \citep{anthropic2026opus5} in Claude Code \citep{anthropic2026claudecode} across all three environments. T2A initialization augments an existing harness with structured execution records needed for diagnosis and gives the development agent access to modify relevant components and their invocation points. Detailed initialization requirements are described in Appendix~\ref{app:baseline-settings}.

For each task agent and environment, every method in Table~\ref{tab:main} is evaluated three times using the same frozen harness. The table reports the means and sample standard deviations of these three evaluations.

\subsection{Overall Performance}
\label{sec:overall}
\input{base/fatal_figure}
Table~\ref{tab:main} compares the resulting frozen systems.

\paragraph{T2A's gains are especially pronounced for task agents and environments with weaker baseline performance.}
With the three environments weighted equally, T2A achieves the highest aggregate score for all three task agents, exceeding the best baseline for Qwen3.5-27B, DeepSeek V4.1 Flash, and Qwen3.5-9B by 12.65, 6.20, and 30.32 points, respectively (Table~\ref{tab:main}, $\Delta$). The initially weakest task agent, Qwen3.5-9B, obtains the largest aggregate gain. For Qwen3.5-27B and DeepSeek V4.1 Flash, the main gains come from ScienceWorld, where baseline performance is relatively weak: mean gains over minimal prompting are 64.6 and 49.1 points, respectively.

\paragraph{The same intervention has different effects across task agents and environments.}
ReAct \citep{yao2023react} improves DeepSeek V4.1 Flash's ScienceWorld mean but lowers both Qwen task agents' means and increases their fatal-failure rates. WorldEvolver \citep{zhang2026worldevolver} improves ScienceWorld means for all three task agents but falls below minimal prompting in ALFWorld for each. These differences show that the same intervention does not guarantee consistent improvement. Section~\ref{sec:deficiencies} uses failure distributions and records of diagnosis, revision, and validation to examine problems across settings and successive harness versions, together with the corresponding revision targets.

\paragraph{A higher mean score does not imply that all current capability bottlenecks have been alleviated.}
WorldEvolver improves the ScienceWorld means of Qwen3.5-27B and DeepSeek V4.1 Flash while increasing their fatal-failure rates. For Qwen3.5-27B, this rate rises from 27.9\% to 30.3\%; T2A instead reduces it to 0.1\% while improving the mean (Figure~\ref{fig:fatal}). However, these aggregate metrics cannot determine whether a revision alleviates its targeted problem. Section~\ref{sec:ablation} uses component ablations to examine changes in scores and target errors; Appendices~\ref{app:score-decomposition} and~\ref{app:rewardmask} provide score decompositions and cases where later failures obscure local progress, respectively.

\subsection{Analysis}
\label{sec:analysis}
\subsubsection{Identifying and Addressing Capability Bottlenecks}
\label{sec:deficiencies}

\begin{figure*}[t]
\centering
\includegraphics[width=\textwidth]{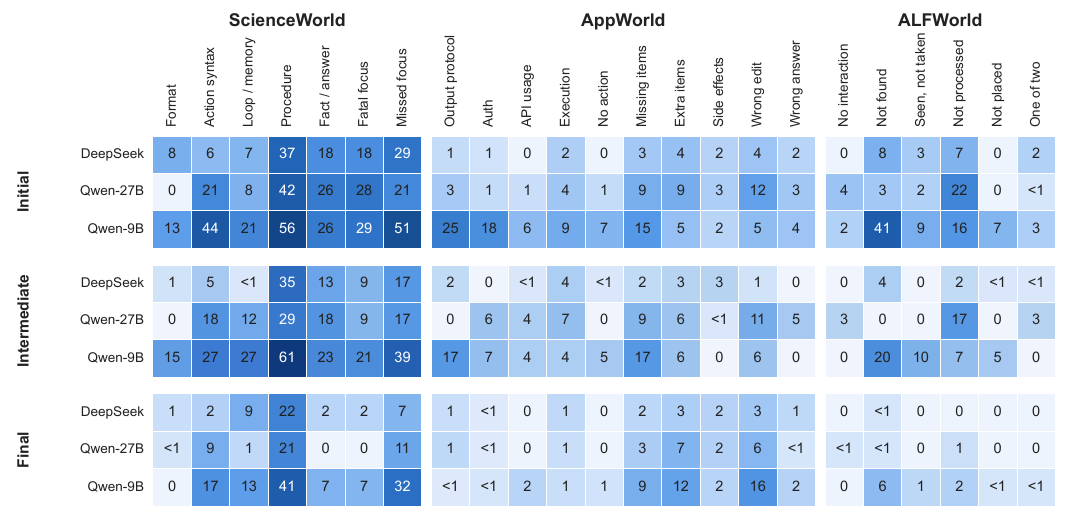}
\caption{Failure categories for initial (minimal-prompt), intermediate, and final T2A harnesses, from one evaluation per harness. Cells show percentages of all test tasks in each environment; $<1$ denotes less than 1\%. Definitions and counting rules are in Appendix~\ref{app:category-definitions}.}
\label{fig:deficiencies}
\end{figure*}

We evaluate the minimal-prompt starting harness, an intermediate harness version, and the final T2A harness once each on the same test tasks within each environment, and classify failures from their execution trajectories. The intermediate version is the accepted harness snapshot nearest the middle of each harness revision sequence. Figure~\ref{fig:deficiencies} shows failure distributions across task agents and environments and how they change across harness versions. Classification and counting rules are in Appendix~\ref{app:category-definitions}.

\paragraph{Different failure patterns call for different revision targets.}
Under minimal prompting, Qwen3.5-9B frequently encounters action-syntax errors in ScienceWorld and output-protocol or authentication failures in AppWorld. DeepSeek V4.1 Flash has fewer such failures, with ScienceWorld losses concentrated more in task procedure and answer errors. For Qwen3.5-9B in ALFWorld, finding the target object is a major obstacle. The records show revisions to output-format and action-syntax handling for Qwen3.5-9B in ScienceWorld, refined planting and life-stage experience for Qwen3.5-27B, and action checks for DeepSeek V4.1 Flash's incorrect focus actions.

\paragraph{Dominant failure types change during revision, prompting renewed diagnosis of capability bottlenecks.}
Although overall task scores improve across the selected snapshots, error categories do not change in parallel. In ScienceWorld, Qwen3.5-9B's action-syntax errors decline across the three snapshots, but procedural errors remain prominent in the intermediate version and decrease only in the final version. The Qwen3.5-27B task agent shows a similar pattern: fatal focus errors decline, whereas loop and memory errors increase in the intermediate version before falling. Paired task analysis in AppWorld shows that, of 384 tasks where Qwen3.5-9B was initially blocked by problems in generating or executing application calls, 216 succeed under the final harness, 150 shift to content-level failures, and only 18 remain in the original categories. As earlier obstacles diminish, problems in object selection and modification content become more prominent. Aggregate task scores alone therefore cannot establish whether a revision alleviates its targeted problem or identify other bottlenecks exposed as earlier obstacles diminish. Appendix~\ref{app:alf-successive} provides a sequence of harness revisions from ALFWorld--DeepSeek V4.1 Flash, showing how T2A uses failures remaining after revision to diagnose bottlenecks anew and adjust subsequent revisions and validation.

\subsubsection{Effects of Targeted Revisions on Capability Bottlenecks}
\label{sec:ablation}

\input{base/component_ablation_table}

In the ScienceWorld--DeepSeek V4.1 Flash system reported in Table~\ref{tab:main}, skill cards, action checks, and within-episode memory address fact/answer errors, erroneous focus actions, and repeated observations, respectively. We remove these components individually and in combination, retaining the other components and basic execution functions, and compare task scores on the same 241 test tasks. Individual-component ablations also examine changes in the corresponding target errors (Appendix~\ref{app:ablation-details}, Table~\ref{tab:target-errors}). Table~\ref{tab:component-ablations} reports the complete score results; evaluation settings and target-error definitions are in Appendix~\ref{app:ablation-details}.

\paragraph{The components selected through diagnosis alleviate their targeted capability bottlenecks.}
Removing skill cards, action checks, and within-episode memory raises the respective counts of tasks with fact/answer errors, focus actions that the rules would block, and repeated observations from 4, 0, and 14 to 28, 30, and 48 (Appendix~\ref{app:ablation-details}, Table~\ref{tab:target-errors}). Combined removals cause larger performance losses than the corresponding individual removals, indicating that final performance depends on contributions from multiple revisions (Table~\ref{tab:component-ablations}). Each component contributes alongside the other revisions; retaining any single component does not preserve the full system's performance.

Further trajectory checks show that all newly fatal tasks after card or action-check removal contain the corresponding target error, whereas this is not the case after memory removal. Trajectories are therefore needed to examine the link between target errors and final score losses; a score decrease alone does not identify its specific cause (Appendix~\ref{app:ablation-details}).

\subsubsection{Starting Harnesses, Instructions, and Workflow}
\label{sec:protocol}
\begin{figure}[t]
\centering
\includegraphics[width=\columnwidth]{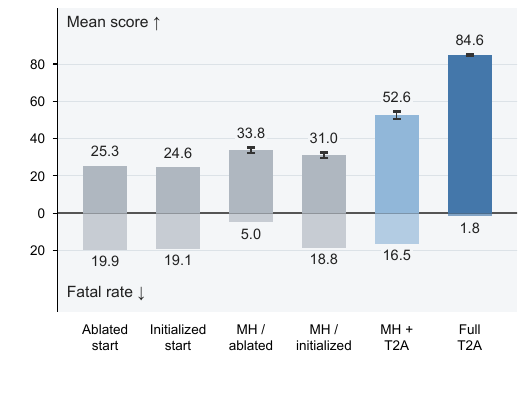}
\caption{ScienceWorld development comparisons with DeepSeek V4.1 Flash. Mean scores appear above zero and fatal-failure rates (\%) below. Results are averaged over three independent development runs; error bars show sample standard deviations of scores. Starting harnesses are evaluated once. MH: Meta-Harness.}
\label{fig:protocol}
\end{figure}

We conduct ablation experiments on ScienceWorld using DeepSeek V4.1 Flash as the task agent. Under a shared development budget, we use Meta-Harness as the base development workflow to examine the effects of individual design choices in T2A. We call the harness augmented to meet T2A's execution-record requirements the initialized start and construct an ablated start from it. To examine the effects of support functions provided at initialization, we remove the record-keeping and execution-support functions listed in Appendix~\ref{app:protocol}, retaining basic execution functions and trajectories. Both conditions use the same development tasks, development model, and resource limits, with three independent development runs per condition. Each resulting frozen harness is evaluated once on the test tasks in Table~\ref{tab:main}. Detailed settings are in Appendix~\ref{app:protocol}.

Meta-Harness improves performance from both starting harnesses, but the initialized start does not yield a higher final mean (Figure~\ref{fig:protocol}). In this comparison, providing these support functions alone is therefore insufficient to improve development outcomes.

\paragraph{T2A's revision instructions improve results under the same conditions.}
Holding the initialized start, accessible records, development model, resource limits, and Meta-Harness's outer evaluation and selection procedure fixed, we replace only the original revision instructions with T2A instructions. These instructions require the development agent to group trajectories by observable features, diagnose problems in light of the distribution of capability bottlenecks, and identify the components or invocation conditions to revise. Before testing, the development agent must specify the expected improvements on target tasks, the performance to preserve on protected tasks, and the evidence needed to establish that the edit was actually triggered. The resulting harnesses' mean score increases by 21.6 points, and all three development runs score above all three runs with the original instructions. This supports explicitly organizing trajectory diagnosis and revision validation.

\paragraph{Full T2A uses diagnosis to guide subsequent revisions.}
Meta-Harness with T2A instructions opens a fresh session each round, with the outer program evaluating and selecting candidates. Full T2A preserves continuous context and lets the development agent select validation tasks based on diagnosis and continue revising, increasing the mean score from $52.6\pm1.9$ to $84.6\pm0.5$. This comparison evaluates the overall development setup, including differences in Claude Code version, reasoning effort, and tool permissions; the gain cannot be attributed solely to context continuity or validation-task selection (Appendix~\ref{app:protocol}).

%% file: base/main_table.tex
\begin{table*}[t]
\definecolor{BestBaseline}{RGB}{229,225,242}
\definecolor{Ours}{RGB}{216,235,232}
\definecolor{T2ABlue}{RGB}{225,238,250}
\definecolor{ModelRed}{RGB}{249,229,231}
\definecolor{Gain}{RGB}{37,105,93}
\small
\centering
\setlength{\tabcolsep}{3pt}
\renewcommand{\arraystretch}{1.22}
\newcommand{\val}[2]{\ensuremath{#1\,{\scriptstyle\pm #2}}}
\resizebox{\textwidth}{!}{%
\begin{tabular}{l|cccccc|>{\columncolor{T2ABlue}}c|>{\columncolor{Ours}}r}
\toprule
\multicolumn{1}{l|}{\multirow{2}{*}[-3pt]{\textbf{Benchmark}}} & \multicolumn{6}{c|}{\textbf{Baselines}} & \cellcolor{white}\textbf{Ours} & \multicolumn{1}{c}{\multirow{2}{*}[-3pt]{$\boldsymbol{\Delta}$}} \\
\cmidrule(lr){2-7}\cmidrule(lr){8-8}
 & Minimal prompt & ReAct & SkillNet & GEPA & WorldEvolver & Meta-Harness & \cellcolor{white}\textbf{T2A} & \multicolumn{1}{c}{} \\
\midrule\rowcolor{ModelRed}\multicolumn{9}{c}{\cellcolor{ModelRed}\textbf{Qwen3.5-27B}}\\\midrule
ScienceWorld & \val{23.1}{1.6} & \val{11.0}{1.1} & \val{31.1}{1.3} & \val{58.6}{0.8} & \val{29.5}{1.8} & \val{63.7}{0.5} & \val{\mathbf{87.7}}{0.1} & \textcolor{black}{$+24.0$} \\
AppWorld Normal & \val{58.9}{2.1} & \val{33.3}{2.7} & -- & \val{86.1}{1.7} & \val{57.1}{1.8} & \val{79.8}{1.2} & \val{\mathbf{90.1}}{1.2} & \textcolor{black}{$+4.0$} \\
AppWorld Challenge & \val{51.9}{0.5} & \val{18.9}{1.1} & -- & \val{72.2}{1.6} & \val{47.5}{0.7} & \val{65.9}{0.7} & \val{\mathbf{74.0}}{1.5} & \textcolor{black}{$+1.8$} \\
ALFWorld Seen & \val{65.2}{1.8} & \val{67.9}{1.4} & \val{89.3}{0.7} & \val{90.0}{0.7} & \val{64.3}{1.4} & \val{91.7}{0.8} & \val{\mathbf{97.4}}{0.4} & \textcolor{black}{$+5.7$} \\
ALFWorld Unseen & \val{69.7}{2.6} & \val{67.2}{2.2} & \val{88.1}{0.7} & \val{94.8}{0.7} & \val{67.9}{2.2} & \val{93.5}{0.4} & \val{\mathbf{99.3}}{0.7} & \textcolor{black}{$+4.5$} \\
\midrule
\textit{Average} & 48.65 & 34.88 & -- & 76.72 & 49.30 & 76.38 & \cellcolor{T2ABlue}\textbf{89.37} & \textcolor{black}{$+12.65$} \\
\midrule\rowcolor{ModelRed}\multicolumn{9}{c}{\cellcolor{ModelRed}\textbf{DeepSeek V4.1 Flash}}\\\midrule
ScienceWorld & \val{36.2}{2.0} & \val{54.8}{1.4} & \val{58.1}{1.4} & \val{64.8}{0.9} & \val{41.3}{1.7} & \val{66.6}{0.7} & \val{\mathbf{85.3}}{0.6} & \textcolor{black}{$+18.7$} \\
AppWorld Normal & \val{82.7}{1.6} & \val{89.9}{1.2} & -- & \val{88.7}{1.0} & \val{72.0}{1.8} & \val{91.5}{0.9} & \val{\mathbf{92.5}}{1.2} & \textcolor{black}{$+1.0$} \\
AppWorld Challenge & \val{78.8}{0.9} & \val{84.4}{1.4} & -- & \val{\mathbf{86.3}}{0.7} & \val{76.5}{0.9} & \val{83.8}{0.3} & \val{84.3}{0.3} & \textcolor{black}{$-2.0$} \\
ALFWorld Seen & \val{78.6}{2.6} & \val{77.9}{1.4} & \val{97.9}{0.0} & \val{87.9}{0.7} & \val{61.4}{2.5} & \val{\mathbf{99.8}}{0.4} & \val{98.1}{0.4} & \textcolor{black}{$-1.7$} \\
ALFWorld Unseen & \val{77.4}{1.6} & \val{76.9}{2.0} & \val{97.0}{0.0} & \val{90.3}{0.7} & \val{61.9}{2.0} & \val{\mathbf{99.0}}{0.4} & \val{\mathbf{99.0}}{0.4} & \textcolor{black}{$0.0$} \\
\midrule
\textit{Average} & 64.98 & 73.12 & -- & 80.47 & 59.07 & 84.55 & \cellcolor{T2ABlue}\textbf{90.75} & \textcolor{black}{$+6.20$} \\
\midrule\rowcolor{ModelRed}\multicolumn{9}{c}{\cellcolor{ModelRed}\textbf{Qwen3.5-9B}}\\\midrule
ScienceWorld & \val{-2.6}{3.1} & \val{-51.6}{4.9} & \val{-16.9}{2.0} & \val{22.4}{1.1} & \val{6.7}{2.9} & \val{39.3}{1.1} & \val{\mathbf{58.3}}{0.9} & \textcolor{black}{$+19.0$} \\
AppWorld Normal & \val{4.2}{1.2} & \val{18.5}{1.0} & -- & \val{33.3}{1.0} & \val{5.4}{1.0} & \val{11.3}{1.0} & \val{\mathbf{69.4}}{1.5} & \textcolor{black}{$+36.1$} \\
AppWorld Challenge & \val{3.7}{0.4} & \val{9.4}{0.7} & -- & \val{22.8}{0.7} & \val{2.9}{0.6} & \val{7.7}{0.6} & \val{\mathbf{50.3}}{1.0} & \textcolor{black}{$+27.5$} \\
ALFWorld Seen & \val{20.5}{2.7} & \val{25.7}{1.9} & \val{62.9}{1.2} & \val{62.9}{1.4} & \val{17.1}{2.6} & \val{66.7}{1.1} & \val{\mathbf{89.0}}{1.1} & \textcolor{black}{$+22.3$} \\
ALFWorld Unseen & \val{23.6}{3.0} & \val{18.7}{2.0} & \val{67.9}{0.7} & \val{69.4}{1.3} & \val{16.4}{2.2} & \val{68.9}{1.1} & \val{\mathbf{89.8}}{0.9} & \textcolor{black}{$+20.4$} \\
\midrule
\textit{Average} & 7.80 & -5.15 & -- & 38.87 & 9.20 & 38.87 & \cellcolor{T2ABlue}\textbf{69.18} & \textcolor{black}{$+30.32$} \\
\bottomrule
\end{tabular}}

\caption{Test results (mean $\pm$ sample standard deviation over three evaluations). Bold marks the highest mean in each row. Average weights the three environments equally, averaging the two splits within AppWorld and ALFWorld. $\Delta$ is T2A's difference from the highest baseline mean in the same row and model block. SkillNet has no AppWorld results or Average because its public release provides no AppWorld skill library or evaluation setup.}
\label{tab:main}
\end{table*}

%% file: base/fatal_figure.tex
\begin{figure*}[t]
\centering
\includegraphics[width=\textwidth]{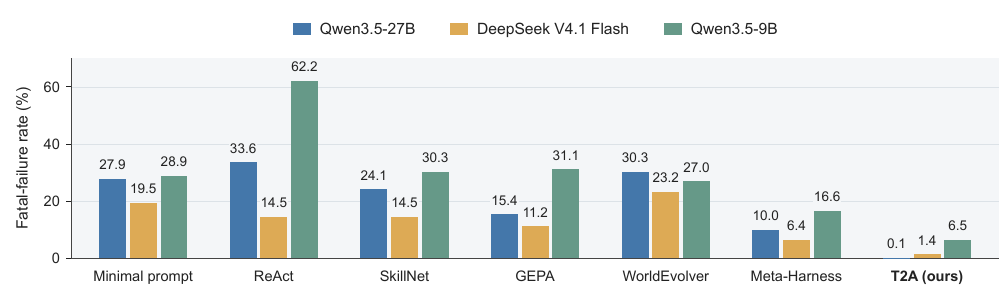}
\caption{ScienceWorld fatal-failure rates, averaged over three evaluations. Lower is better; colors distinguish the three task models.}
\label{fig:fatal}
\end{figure*}

%% file: base/component_ablation_table.tex
\begin{table}[t]
\centering
\small
\renewcommand{\arraystretch}{1.12}
\setlength{\tabcolsep}{8pt}
\begin{tabular}{@{}lrr@{}}
\toprule
\textbf{Removed} & \textbf{Mean} & $\boldsymbol{\Delta}$ \\
\midrule
None (full) & \textbf{85.2} & --- \\
\midrule
Cards & 65.0 & $-20.2$ \\
Check & 70.4 & $-14.8$ \\
Memory & 70.9 & $-14.3$ \\
\addlinespace[3pt]
Cards + check & 52.3 & $-32.9$ \\
Cards + memory & 46.5 & $-38.7$ \\
Check + memory & 57.4 & $-27.8$ \\
\addlinespace[3pt]
All components & 37.7 & $-47.5$ \\
\bottomrule
\end{tabular}
\caption{Individual and combined component ablations on ScienceWorld. $\Delta$ is the task-paired score difference from the full system. Evaluation and implementation details are in Appendix~\ref{app:ablation-details}.}
\label{tab:component-ablations}
\end{table}

%% file: base/5_conclusion.tex
\section{Conclusion}
We presented T2A, which uses execution evidence to diagnose failures, select harness revisions, and validate their effects without updating model weights. Across three environments and three task agents, T2A achieves the highest aggregate score among the evaluated methods. With the starting harness and Meta-Harness outer loop fixed, three independent development runs per condition show that T2A revision instructions improve the mean ScienceWorld score by 21.6 points. Component ablations link performance gains to reductions in target errors, while development records show how remaining failures guide subsequent revisions. These findings support the view that different task agents and harnesses face different capability bottlenecks, and that the effects of the same intervention vary accordingly. Revision targets should therefore be chosen based on the distribution of deficiencies in the current harness, and checking their effects on capability bottlenecks should guide further revisions.\label{sec:conclusion-end}

%% file: base/6_limitations.tex
\section*{Limitations}
T2A depends on the development agent's trajectory analysis, interpretations, and implementation choices. Results may vary with its underlying model, retained context, and tools. Updates to Claude Code may change context management, tool use, and execution behavior, so reproducing the harness revisions requires checking its version and configuration.

We ablate only components that can be independently disabled. Other revisions are coupled with the basic execution flow: directly removing them would disrupt normal operation, while replacing them would require corresponding changes to the execution logic. They are therefore excluded from the component ablations in this study.

%% file: base/appendix.tex
\appendix
\raggedbottom
\makeatletter
\setlength{\@fptop}{0pt}
\setlength{\@dblfptop}{0pt}
\setlength{\@fpsep}{16pt}
\setlength{\@dblfpsep}{16pt}
\makeatother
\setlength{\floatsep}{12pt plus 2pt minus 2pt}
\setlength{\textfloatsep}{14pt plus 2pt minus 2pt}
\setlength{\intextsep}{12pt plus 2pt minus 2pt}
\renewcommand{\arraystretch}{1.12}
\setlength{\LTcapwidth}{\textwidth}
\setlength{\LTpre}{10pt}
\setlength{\LTpost}{10pt}
\setcounter{dbltopnumber}{3}
\renewcommand{\dbltopfraction}{0.9}
\renewcommand{\topfraction}{0.9}
\renewcommand{\bottomfraction}{0.8}
\renewcommand{\textfraction}{0.08}
\renewcommand{\floatpagefraction}{0.75}
\renewcommand{\theequation}{\thesection\arabic{equation}}

\input{base/appendices/Q_related_comparison}
\FloatBarrier
\appsec{T2A Workflow}{app:t2a-workflow}
Algorithm~\ref{alg:t2a} summarizes Section~\ref{sec:method}. Each round groups trajectories below full score, $\mathcal F_r$, by observable features into $\mathcal G_r$. Candidate diagnoses $\mathcal Z_r^{\mathrm{cand}}$ are checked against related successful and failed trajectories to form $\mathcal Z_r$ (Equation~\ref{eq:diagnosis-record}). Groups organize investigation; diagnoses and their applicability conditions guide the edit $\delta$. Every selected task is diagnosed or marked unresolved.

The plan $\Pi_r$ records target outcomes, protected tasks, edit-triggering checks, and conditions $\mathcal V_r$. From current and candidate trajectories, $\mathcal T_r^{\mathrm{val}}$ and $\mathcal T'_r$, the development agent assigns each condition $1$ when met and $0$ otherwise; insufficient records count as unmet. Acceptance ($a_r=1$) requires all conditions to hold. Otherwise, $H_r$ remains accepted while the candidate is revised, revalidated, or withdrawn.
\input{base/t2a_algorithm}
\FloatBarrier
\input{base/appendices/I_M_method}
\input{base/appendices/N_experiments}
\input{base/appendices/O_cases}

\appsec{Use of AI Assistants}{app:ai-use}
We used AI assistants for code implementation and debugging, literature search, analysis support, and manuscript drafting, revision, and translation. The authors are responsible for the final manuscript, code, and reported results.

%% file: base/appendices/Q_related_comparison.tex
\appsec{Comparison of Revision Procedures}{app:revision-comparison}
Table~\ref{tab:revision-comparison} compares how related methods turn execution feedback into harness edits and decide which edits to retain. Several methods already use failure diagnosis, scoped edits, and regression checks. The relevant distinction is therefore how evidence is organized, how it determines the scope of a change, and how acceptance is decided.

\begin{table*}[t]
\centering
\small
\begingroup
\setlength{\tabcolsep}{5pt}
\renewcommand{\arraystretch}{1.18}
\begin{tabularx}{\textwidth}{>{\raggedright\arraybackslash}p{0.16\textwidth}YYY}
\toprule
\textbf{Method} & \textbf{Diagnostic evidence} & \textbf{How edits are organized} & \textbf{Validation and selection} \\
\midrule
\rowcolor{black!6}
Self-Harness \citep{zhang2026selfharness} & Failed trajectories clustered by verifier-grounded failure mechanisms. & Bounded component edits with audit records and passing behaviors to preserve. & Held-in and held-out pass counts must not decrease; at least one must improve. \\
\addlinespace[3pt]
AHE \citep{lin2026ahe} & Layered trajectory reports and manifests linking evidence, causes, and predicted effects. & Component edits in a decoupled harness workspace. & Next-round attribution checks scores for expected fixes and regressions and supports rollback. \\
\addlinespace[3pt]
\rowcolor{black!6}
HarnessX \citep{chen2026harnessx} & Task briefs and a cross-task outcome overview with edit history. & Edits to typed processors, recorded in change manifests. & A predefined procedure for review, checks, and regression gates. \\
\addlinespace[3pt]
HarnessFix \citep{chen2026harnessfix} & Trace intermediate representation links responsible steps to harness-layer flaw records. & Repair operators instantiated with file and behavior constraints. & Repair-scope checks, target-flaw reduction, and regression validation. \\
\addlinespace[3pt]
\rowcolor{black!6}
EvolveNet \citep{nie2026evolvenet} & Client-specific experience and evidence of gains and regressions. & Aggregate local program adaptations into shared or domain-specific changes. & Commit the integrated harness only if validation gains at least match regressions. \\
Meta-Harness \citep{lee2026metaharness} & The proposing agent reads historical code, scores, and execution trajectories. & Free-form code changes chosen by the proposer. & Search-set evaluation and a performance--cost Pareto frontier. \\
\addlinespace[3pt]
\midrule
\textbf{T2A (ours)} & Structured execution records provide failure features and error clusters. Hypotheses from representative trajectories are checked against related trajectories to establish concrete execution evidence for locating capability bottlenecks. & Bottleneck diagnosis and test feedback determine changes to content, invocation, or execution, and the scope of those changes. & Guided by bottleneck diagnosis, the development agent judges whether preregistered target-task, protected-task, and edit-triggering conditions are met. \\
\bottomrule
\end{tabularx}
\caption{How related methods diagnose failures, organize edits, and validate candidates.}
\label{tab:revision-comparison}
\par\smallskip
\endgroup
\end{table*}

\paragraph{Relation to the other procedures.}
The comparison locates T2A's contribution in organizing diagnosis and validation. It makes the scope of a failure hypothesis explicit before selecting an edit, and keeps unsupported cases unresolved. This complements free-form proposer search, failure-mechanism mining, edit-effect attribution, layer-specific repair, and collaborative program aggregation. Considering the applicability of these procedures to the selected ScienceWorld task set, Section~\ref{sec:protocol} evaluates T2A development instructions within a fixed Meta-Harness outer loop.

%% file: base/t2a_algorithm.tex
\begin{algorithm}[tbp]
\caption{T2A Workflow}
\label{alg:t2a}
\footnotesize
\begin{algorithmic}[1]
\Require Initial harness $H_0$; development tasks $\mathcal Q_r$
\Ensure Accepted harness
\For{each round $r$ within the experiment limits}
\Statex \textit{Data collection}
\State $\mathcal T_r\gets\Call{Collect}{H_r,\mathcal Q_r}$ \Comment{Eq.~\ref{eq:traces}}
\Statex \textit{Harness revision}
\State $\mathcal F_r\gets\{\tau_d\in\mathcal T_r:s_d<s_d^{\max}\}$
\State $\mathcal G_r\gets\Call{Group}{\mathcal F_r}$
\State $\mathcal Z_r^{\mathrm{cand}}\gets\Call{Explain}{\mathcal G_r,H_r}$
\State $\mathcal Z_r\gets\Call{Check}{\mathcal Z_r^{\mathrm{cand}},\mathcal T_r,H_r}$
\Repeat
\State $\delta\gets\Call{Revise}{H_r,\mathcal Z_r}$ or retain $\delta$
\State $H'\gets\delta(H_r)$
\State $\Pi_r\gets\Call{Register}{H',H_r}$
\State $(\mathcal T_r^{\mathrm{val}},\mathcal T'_r)\gets\Call{Validate}{H',H_r,\Pi_r}$
\State $a_r\gets\bigwedge_{v\in\mathcal V_r}v(\mathcal T'_r,\mathcal T_r^{\mathrm{val}})$
\State Retain supported findings for further revision
\State If $a_r=0$, revise, revalidate, or withdraw $H'$
\Until{$a_r=1$ or $H'$ is withdrawn}
\State Set $H_{r+1}\gets H'$ if $a_r=1$; otherwise retain $H_r$
\EndFor
\State \Return Latest accepted harness
\end{algorithmic}
\end{algorithm}

%% file: base/appendices/I_M_method.tex
\appsec{Conditional-Score Decomposition}{app:score-decomposition}
Let $B$ denote the target behavior, $\bar B$ its absence, and $J(H)=E_H[s]$ the expected terminal score over tasks and execution randomness under harness $H$.
Let $p_r=P_r(B)$ and $p'=P'(B)$ denote the target behavior's probability under the current and candidate harnesses. Write $\mu_r^+=E_r[s\mid B]$ and $\mu_r^-=E_r[s\mid\bar B]$, with analogous definitions under the candidate. These conditional expectations require positive probability for the conditioning events. With $\Delta p=p'-p_r$ and $\Delta\mu^{\pm}=\mu'^{\pm}-\mu_r^{\pm}$, the law of total expectation gives
\begin{equation}
J(H_r)=p_r\mu_r^+ +(1-p_r)\mu_r^-.
\label{eq:events}
\end{equation}
Subtracting the current expectation from the candidate expectation gives $\Delta J=J(H')-J(H_r)$:
\begin{equation}
\Delta J=\Delta p\big(E_r[s\mid B]-E_r[s\mid\bar B]\big)+\Delta_{\mathrm{other}},
\label{eq:decomposition}
\end{equation}
where
\begin{equation}
\Delta_{\mathrm{other}}=p'\Delta\mu^+ +(1-p')\Delta\mu^-.
\label{eq:remainder}
\end{equation}
The remainder includes changes in subsequent execution and in the composition of the samples conditioned on the target behavior.

%% file: base/appendices/N_experiments.tex
\appsec{Experimental Settings, Additional Results, and Trajectory Analysis}{app:experiments}
\subsection{Harness Revision Requirements and Baseline Adaptations}
\label{app:baseline-settings}
\paragraph{Licensing.}
Our original code is released under the MIT License; third-party materials retain their respective licenses.

\paragraph{T2A initialization and environment adaptation.}
T2A initialization augments an existing harness with structured execution records needed for diagnosis and gives the development agent access to modify relevant components and their invocation points. Records should be linked by task, execution step, and model-call identifiers so that the developer can trace the input actually received by the model and its sources, the model's reply, parsing and action processing, and the action ultimately submitted and its environment feedback. For existing retrieval, memory, or action-checking components, records should also capture invocation outcomes and whether relevant content entered the model input, distinguishing missing content, absent invocation, failed delivery, and noncompliance after delivery.

Appendix~\ref{app:protocol} details the ScienceWorld starting harness used in the controlled comparison.

\paragraph{Initial support functions.}
Table~\ref{tab:initial-support} summarizes the recording and execution functions present in each experimental starting harness.

\begin{table*}[t]
\centering\small
\begingroup
\setlength{\tabcolsep}{6pt}
\renewcommand{\arraystretch}{1.22}
\begin{tabularx}{\textwidth}{>{\raggedright\arraybackslash}p{3.6cm}YYY}
\toprule
\textbf{Function} & \textbf{ScienceWorld} & \textbf{AppWorld} & \textbf{ALFWorld}\\\midrule
Structured records & Present & Present & Present\\
\rowcolor{black!6}
Within-episode memory & Interaction history; confirmed object locations & Interaction history & Recent history\\
Experience retrieval & Empty card library & None & Empty card library\\
\rowcolor{black!6}
Action processing & On & None & None\\
\shortstack[l]{Action checks /\\execution repair} & Off & None & None\\\bottomrule
\end{tabularx}
\endgroup
\caption{Support functions in the initial harnesses. Off denotes code that is present but disabled, and none denotes an absent component.}
\label{tab:initial-support}
\end{table*}

\emph{Structured records} are per-step records linked by task, step, and model-call identifiers, covering model input and its sources, the reply, parsing and action processing, the submitted action, and environment feedback. \emph{Within-episode memory} is information carried across steps and included in model input. In ScienceWorld, it includes interaction history and object locations confirmed by environment receipts; AppWorld uses interaction history, and ALFWorld uses recent history. \emph{Experience retrieval} places stored experience cards in model input; an empty library retains the mechanism but contains no cards. \emph{Action processing} rewrites or completes a proposed action before execution. In the ScienceWorld start, this includes matching container names to objects in the latest observation and completing mixing steps when the recipe is fully public. \emph{Action checks / execution repair} reject proposed actions with feedback before execution or repair failed execution. These functions are distinct from action processing.

To provide intervention points for guidance and structured action processing, initialization adjusts the flow from model output to environment execution. At the starting stage, some interfaces are not yet equipped with the guidance and checking functions to be developed later, so these flow changes may reduce initial task performance. To maintain comparability of the development interfaces and editable scope, the ablated start retains these flow changes while removing only the additional structured records, object-location memory, and action-processing functions, preserving basic execution functions and action--observation--score trajectories.

\paragraph{WorldEvolver.}
We adapt WorldEvolver \citep{zhang2026worldevolver} to build memory on development tasks and freeze it for testing. For comparability with the other methods in Table~\ref{tab:main}, repeated attempts and memory updates are restricted to training tasks; each test task is attempted once per evaluation. The ALFWorld adaptation uses current observations and does not allow queries for admissible-action lists that are not directly visible to the player.

\subsection{Experimental Setup and Task Splits}
\label{app:development}
\paragraph{Model execution and scoring.}
Qwen models \citep{qwen2026modelcards} use local inference services in all three environments; DeepSeek V4.1 Flash \citep{deepseek2026v41} is accessed through the official API. The development and reflection model is Claude Opus 5 \citep{anthropic2026opus5}. ScienceWorld retains the last environment score when an output limit or a terminal condition is reached, and reports mean terminal score and fatal-failure rate. ALFWorld reports task completion percentages on \texttt{valid\_seen} and \texttt{valid\_unseen}; AppWorld reports task completion percentages on \texttt{test\_normal} and \texttt{test\_challenge}.

\paragraph{Software and developer settings.}
ALFWorld uses version 0.4.2 and AppWorld uses 0.1.3.post1. ScienceWorld uses 1.2.2 for T2A and GEPA and 1.2.3 for minimal prompting, ReAct, and SkillNet. The Meta-Harness baseline and all Meta-Harness arms in Section~\ref{sec:protocol} use Claude Code 2.1.285 with model identifier \texttt{claude-opus-5} and \texttt{--effort max}. The ScienceWorld Full T2A launcher uses Claude Code 2.1.284 with the same model identifier and default effort. Its session runs in noninteractive mode, with web access, MCP, and subagents disabled; it uses \texttt{--max-turns 1000} and resumes interrupted work in the same session. These launch commands do not explicitly override the developer model's temperature or top-$p$. The released ALFWorld and AppWorld T2A files do not pin a developer CLI version or effort setting.

\paragraph{T2A task-model decoding and limits.}
Thinking is disabled for the task models. In ALFWorld, all task models have a 100-interaction limit. The Qwen models use temperature 0.7, top-$p$ 0.8, top-$k$ 20, and presence penalty 1.5; the per-call output limits are 1,024 tokens for Qwen3.5-27B and 128 for Qwen3.5-9B. DeepSeek V4.1 Flash uses temperature 0.5 and a 512-token output limit. ScienceWorld uses temperature 0.5, at most 1,200 output tokens per call, 100 environment interactions, and 180 model calls per episode. In AppWorld, the Qwen models use temperature 0, top-$p$ 0.8, top-$k$ 20, presence penalty 1.5, and a 2,048-token output limit; DeepSeek V4.1 Flash uses temperature 0.5 and a 4,096-token output limit. T2A allows 40 turns per AppWorld task; the ReAct baseline retains its upstream 50-turn limit.

\paragraph{Task splits.}
For the experiments reported in Table~\ref{tab:main}, ScienceWorld uses 178 tasks from \texttt{train} for harness revision, selecting seven tasks from each of the 26 task types, or all available tasks when a type has fewer than seven. Its 241-task evaluation subset is selected from \texttt{test} across 26 task types: within each type, we sort by variation ID and take the first 10 variants, or all variants when fewer than 10 are available \citep{wang2022scienceworld}.

ALFWorld uses 138 tasks from \texttt{train} for harness revision, with 23 tasks from each of its six task types. Evaluation covers 140 tasks in \texttt{valid\_seen}, set in rooms seen during training, and 134 tasks in \texttt{valid\_unseen}, set in unseen rooms \citep{shridhar2021alfworld}.

AppWorld uses \texttt{train} and \texttt{dev} for harness revision. The 168 tasks in \texttt{test\_normal} use the same application pool as the development splits. The 417 tasks in \texttt{test\_challenge} each require at least one application held out from those splits, Amazon or Gmail \citep{trivedi2024appworld}.

For both the experiments reported in Table~\ref{tab:main} and the comparison of harness revision instructions, test tasks and their evaluation feedback are not used for harness revision or selection. Harnesses are revised separately for each task agent and frozen before testing.

\subsection{Component Ablation Settings}
\label{app:ablation-details}
The ablations use the DeepSeek V4.1 Flash ScienceWorld system reported in Table~\ref{tab:main} on the same 241 test tasks. Each arm removes one or more component groups while keeping the others unchanged. Combined ablations are reimplemented using the same component definitions. Each ablated harness is evaluated once and paired by task with the first evaluation of the full system among those summarized in Table~\ref{tab:main} (mean 85.21), collected in a separate batch. Table~\ref{tab:component-ablations} reports the means and paired score differences. Target-error counts and trajectory checks below concern the individual-component ablations.

Removing the skill-card library disables card retrieval. Removing action checks bypasses the pre-execution checks for invalid or repeated ineffective actions and task-specific focus prerequisites, together with their blocking feedback. Basic action parsing and environment execution remain unchanged. Removing within-episode memory hides the stored observations and progress from model input; action checks still use it internally. Operating rules are not ablated because removing them would disrupt normal execution and require a manually chosen boundary.

Target-error criteria were defined from records collected during harness revision before this ablation evaluation. For cards, whose design targets procedural and factual knowledge, the reported target error is a fact/answer error identified by blind trajectory reading. For the check component, designed to address erroneous commitments and ineffective repetition, the reported criterion is execution of a focus action that the frozen check rules would block. For memory, the criterion is at least three repeated observations whose results are already known. Counts denote episodes containing the specified error. Newly fatal tasks have a nonfatal full-system outcome and a score of $-100$ after component removal. Of the 16 newly fatal tasks after removing the check component, 12 reach $-100$ at the flagged action itself.

\begin{table}[t]
\centering
\small
\begingroup
\setlength{\tabcolsep}{6pt}
\renewcommand{\arraystretch}{1.22}
\begin{tabular}{lrr}
\toprule
\textbf{Component} & \textbf{Full} & \textbf{Removed} \\
\midrule
Cards & 4 & 28 \\
\rowcolor{black!6}
Check & 0 & 30 \\
Memory & 14 & 48 \\
\bottomrule
\end{tabular}
\endgroup
\caption{Tasks containing the target error in the full system and after individual component removal. Each row uses its component-specific error definition described in this section.}
\label{tab:target-errors}
\end{table}

\subsection{Starting Harnesses, Instructions, and Workflow}
\label{app:protocol}

\paragraph{Tasks and starting harnesses.}
The task agent is DeepSeek V4.1 Flash. For comparability, all methods in this comparison receive the same pool of 178 official training tasks spanning 26 types and a budget of 356 development-evaluation episodes after collecting the starting trajectories. This budget excludes the 178 episodes used to collect those trajectories. By comparison, Meta-Harness in Table~\ref{tab:main} executes 1,378 development episodes from the minimal-prompt start and reaches a mean score of 66.6. Frozen harnesses are evaluated on the same 241 ScienceWorld test tasks as in Table~\ref{tab:main}. Training and test tasks are disjoint.

We construct a support-function ablation from the T2A-initialized harness. The initialized start includes the diagnostic records and editable components described in Appendix~\ref{app:baseline-settings}, together with the object-location memory and action-processing functions of this ScienceWorld implementation. The ablated start removes the additional structured records, object-location memory, and action processing while retaining model calls, action parsing, environment interaction, basic within-episode state, and action--observation--score trajectories. Meta-Harness revises both starting harnesses with the same development tasks, development model, and resource limits. This comparison evaluates the joint contribution of these support functions to the performance of the resulting harnesses.

\paragraph{Functions in the initialized start.}
Both starts log episode boundaries and each step's action, observation, and score. The initialized start additionally records model-input components, retrieval candidates, and prompt-token counts per call, together with rejected replies and corrective feedback. Environment-step records include object locations, public facts, the action-processing route, and the action actually submitted. A separate call ledger retains raw model replies; episode summaries record costs and component activity.

Object-location memory stores moves confirmed by environment receipts, including the destination and supporting step. A later action that may move the object without a confirming receipt invalidates the entry. The location table and public facts enter each model prompt.

In chemical and state-change tasks, action processing matches container names to the objects in the latest observation and rewrites the action accordingly. It can also complete mixing steps when the recipe is fully public. Other action-check and execution-repair components are disabled in this start. The skill-card retrieval and delivery mechanism is present with an empty card library. Object-location memory and action processing are features of this experimental starting harness, not requirements for initializing T2A in every environment.

\paragraph{Replacing the harness revision instructions.}
From the same T2A-initialized start, we replace the original Meta-Harness harness revision instructions \citep{lee2026metaharness} with T2A instructions while retaining its outer evaluation and selection procedure and run settings. T2A instructions require the developer to group trajectories, propose and check explanations in context, and validate revisions against predefined conditions. Each condition has three independent development runs.

\paragraph{Full T2A harness revision.}
Full T2A uses the T2A-initialized start in a continuous session with the development agent. It receives starting trajectories for 178 tasks, follows the T2A workflow to select validation tasks and repeat counts, and continues revising the harness based on diagnosis and validation results. Full T2A also has three independent development runs. Unlike the instruction-replacement comparison, which holds the developer settings fixed, the Full T2A comparison also differs in Claude Code version, effort, and tool permissions as described in Appendix~\ref{app:development}. It therefore evaluates the combined development setup.

\paragraph{Selection, stopping, and context.}
In Full T2A, the development agent writes and judges its validation conditions from execution records without researcher review. The program enforces the budget, records workspace fingerprints, and freezes submissions; it does not independently judge whether the semantic acceptance conditions hold. The last explicitly submitted harness is evaluated; if there is no submission, the starting harness is used. Full T2A uses one continuous development session and may finish at the agent's discretion. Exhausting the episode budget prevents further environment runs but still permits review of existing evidence and submission. Context is managed by Claude Code's built-in compaction \citep{anthropic2026codecontext}, without a separate context-recovery mechanism. No separate limit is imposed on inner revision attempts; environment runs remain subject to the total episode budget. No per-round candidate-proposal time limit is imposed. By contrast, all Meta-Harness arms use up to five rounds, each with a fresh session and a 2,400-second proposal limit. A round that produces no candidate is not rerun. The outer program evaluates submitted candidates and selects the highest search-set mean, including the starting harness; ties favor the earlier candidate. This selection rule also applies to Meta-Harness with T2A instructions.

\paragraph{Evaluation settings.}
All arms use DeepSeek V4.1 Flash at temperature 0.5 with thinking disabled, at most 1,200 output tokens per call, 100 environment interactions, and 180 model calls per episode. These task-execution limits are enforced outside the editable harness. Scores are recorded for all 241 tasks. Each of the three harnesses from independent development runs is evaluated once. When the execution budget is exhausted or a terminal condition is reached, the last available environment score is retained.

\subsection{Failure Classification and Trajectory Analysis}
\label{app:deficiency-sources}

\paragraph{Category definitions and denominators.}
\label{app:category-definitions}

Figure~\ref{fig:deficiencies} uses one evaluation per harness. Each cell is a count divided by the size of the environment's whole test set; all three blocks share the same denominator. The intermediate version is the accepted harness snapshot nearest the middle of each harness revision sequence. For ScienceWorld--DeepSeek V4.1 Flash, this is the snapshot at development episode 190.

\textbf{ScienceWorld} ($n=241$ episodes). Each cell counts point-losing episodes ($s<100$) exhibiting the phenomenon; an episode may receive several labels.
\begin{itemize}
\setlength{\itemsep}{2pt}
\item \emph{Format}: at least 10\% of model replies contain no executable action.
\item \emph{Action syntax}: at least 10\% of submitted actions receive ``No known action matches that input.''
\item \emph{Loop / memory}: ten consecutive steps without a score change use at most three distinct actions, none of them \texttt{wait}.
\item \emph{Procedure}: points are lost because required steps are not carried out in a workable order.
\item \emph{Fact / answer}: a committed answer is wrong because of an incorrect fact or identification.
\item \emph{Fatal focus}: final score is $-100$.
\item \emph{Missed focus}: no focus action is accepted during the episode.
\end{itemize}
Procedure and Fact / answer are labelled by Claude Opus 5 \citep{anthropic2026opus5} using anonymized, shuffled trajectories without system identities; the other five categories are computed mechanically.

\textbf{AppWorld} ($n=585$ tasks). Each failed task receives one primary cause. Rules are checked in a fixed order and the first match wins, so an earlier cause can mask a later one.
\begin{itemize}
\setlength{\itemsep}{2pt}
\item \emph{Output protocol}: parser-rejected replies, truncated output, misuse of the answer slot, or results never printed.
\item \emph{Auth}: requests rejected as unauthorized, with no successful app call.
\item \emph{API usage}: turns spent on unknown APIs or invalid arguments.
\item \emph{Execution}: turn limit, identical code resubmitted, context budget, or infrastructure error.
\item \emph{No action}: the required change is never made.
\item \emph{Missing items}, \emph{Extra items}, \emph{Side effects}, and \emph{Wrong edit}: a change is made, but items are omitted, extra items included, unrelated records changed, or a field value or selected object is wrong, respectively.
\item \emph{Wrong answer}: only the returned answer value is wrong.
\end{itemize}

\textbf{ALFWorld} ($n=274$ tasks). For each harness, its trajectories are classified independently: each failed task is assigned to the first task milestone it never reaches.
\begin{itemize}
\setlength{\itemsep}{2pt}
\item \emph{No interaction}: fewer than three steps, or most replies rejected.
\item \emph{Not found}: the target object never appears.
\item \emph{Seen, not taken}: the target appears but is never picked up.
\item \emph{Not processed}: the required clean, heat, cool, or light step is not completed.
\item \emph{Not placed}: the object never reaches the target receptacle.
\item \emph{One of two}: only one object is placed in a two-object task.
\end{itemize}

\paragraph{ReAct on ScienceWorld with Qwen3.5-9B.}
We read paired trajectories from one minimal-prompt evaluation and one ReAct evaluation on the same 241 tasks, using the first evaluation of each harness among those summarized in Table~\ref{tab:main}. Under the definitions above, the percentage of episodes with format errors falls from 13\% to 3\%, while fatal failures rise from 29\% to 62\%. Among the additional failures under ReAct, 34.1\% contain behavior that contradicts descriptions already present in the model's prompt. In 61.9\%, ReAct makes task progress earlier than minimal prompting but later executes a fatal action despite having more steps remaining. These trajectories suggest that improving one kind of execution problem can shift the capability bottleneck: the model makes progress earlier, yet still fails because of a later incorrect decision.

%% file: base/appendices/O_cases.tex
\appsec{Historical Diagnosis and Revision Cases}{app:cases}
\begin{table*}[t]

\small\centering
\begingroup
\setlength{\tabcolsep}{6pt}
\renewcommand{\arraystretch}{1.18}
\begin{tabularx}{\textwidth}{>{\raggedright\arraybackslash}p{2.0cm}YYY}
\toprule
\textbf{Task situation}&\textbf{Old trajectory}&\textbf{Revised trajectory}&\textbf{What the trajectories establish}\\\midrule
Prepare an intermediate color for tertiary mixing&Repeated kitchen search without finding materials; final score 0&Enter the art studio, mix blue/yellow paint; environment confirms green at 23 points; then select the wooden cup as the target and end at $-100$&Material location and preparation improve, but target reference remains faulty; lower final score obscures earlier progress\\[4pt]
\rowcolor{black!6}
Compare inclined-plane angles&Search/navigation without a descent experiment; final score 10&Complete the first descent and read time; while preparing the second, use \texttt{focus} on the stopwatch as a reset and end at $-100$&First experiment progresses, but using focus to reset the stopwatch ends the task\\[4pt]
Infer dominant/recessive plant traits&Sow seeds but fail to obtain water effectively; plants do not grow; final score 16&Grow two parent plants to flowering, but answer from parental leaf shape instead of offspring evidence; end at $-100$&Care operations progress, but genetic evidence remains insufficient\\\bottomrule
\end{tabularx}
\endgroup
\caption{Historical same-task cases with local progress and subsequent fatal failure, evaluated under the execution limits in effect when each harness version was evaluated.}
\label{tab:rewardmask}
\end{table*}

\subsection{Successive Bottlenecks in ALFWorld with DeepSeek V4.1 Flash}
\label{app:alf-successive}
This case follows diagnosis, revision, and validation on ALFWorld training tasks with DeepSeek V4.1 Flash.

\paragraph{From delivered guidance to action checks.}
In one failed task, the goal required a mug, but the task agent took and cooled a cup and failed after 100 steps. The exact-type skill card had been retrieved and included in the prompt. Inspection of 13 failed episodes identified nine in which delivered guidance was not followed. The developer therefore converted the type rule into a pre-execution action check: actions selecting the wrong object type were rejected, and the rejection reason entered the next model call.

A counterexample exposed an error in this check. In a look-at task, a lamp serves as a tool; rejecting its pickup because it does not match the target object's type blocks a valid action. The revised check distinguished task objects, tools, and containers. On the same 120 training tasks, completion increased from 85 under minimal prompting to 98 under the candidate. However, successful two-object tasks fell from 17 to seven, and 17 tasks that minimal prompting completed now failed.

\paragraph{Diagnosing the remaining failures.}
The developer read the 17 regressions and all 22 candidate failures. In 16 regressions, the first error involved an action being allowed despite an unmet observable prerequisite: moving toward a target that was not visible, taking a second object before putting down the first, or retrieving an object already placed correctly. The trajectories did not support the earlier explanation that search feedback caused the regressions, so that explanation was withdrawn.

The next revision added checks for visibility, whether the agent was already holding an object, and whether an object had already been placed correctly. Implementation review also found that opening a container was incorrectly treated as a room change; this state-tracking error was corrected before validation.

\paragraph{Validating behavior and selecting revisions.}
A $2\times2$ comparison independently enabled or disabled the revised action checks and the new card library on 32 training tasks, with two evaluations per condition. Enabling the checks without the new cards increased completion from 21 to 31 in both evaluations. Adding the new cards on top of the checks changed completion from 31 to 31 and from 31 to 30. The checks were accepted, and the new card library was withdrawn.

The trajectories also showed changes in the targeted behavior. Visibility checks rejected 186 actions; in 168 cases, the next action switched to navigation or search. Checks against taking another object while already holding one rejected 59 actions; 35 were followed by putting down the held object. The check against retrieving an already placed object had only one opportunity to intervene; it rejected the action, and that episode succeeded. In a subsequent evaluation on all 120 training tasks, the frozen harness completed 118. This sequence shows how validation exposed a new bottleneck after the first revision and guided the next revision and acceptance decision.

\subsection{Reward Structure Can Obscure Local Progress}
\label{app:rewardmask}
We revisit historical paired outcomes on the same 184 tasks, including 31 transitions from nonfatal endings to fatal failures. After excluding nine tasks that already had full scores under the old harness, we read complete old/new trajectories for the remaining 22, distinguishing operational progress, experimental evidence, and fatal submissions. Table~\ref{tab:rewardmask} gives three representative situations.

\subsection{A Diagnosis-to-Revision Example: Life-Span Ranking}
\label{app:lifespan-revision}
This example summarizes a Full T2A development record in ScienceWorld with DeepSeek V4.1 Flash, showing a shift from skill-card content to control logic.

\paragraph{Triage and explanation checks.}
A script selected 118 point-losing trajectories from the 178 starting trajectories. Of the 27 trajectories ending at $-100$, 24 ended with a \texttt{focus on} action. This pattern motivated the explanation that focus submits an answer and that selecting a present but incorrect object can cause fatal failure. Counterexamples narrowed the explanation: in \texttt{measure-melting-point-}\allowbreak\texttt{known-substance-v0157}, two focuses on the same object at steps 12 and 14 were nonfatal; in \path{use-thermometer-v0125}, a focus on an object absent from the room at step 36 was rejected by the parser without fatal failure. Thus neither repeated focus nor parser rejection alone implied a fatal outcome.

\paragraph{Card revision and regression.}
The first revision added life-span cards with typical species life spans and stated that an egg counts as an animal of its species. Two target tasks improved from $-100$ to 100, but the protected task \path{lifespan-shortest-lived-v0056} fell from 100 to $-100$: the model chose a frog egg (six years) over a baby mouse (four years). The developer neutralized the wording and added a worked example. In a later round, the same task again ended at $-100$ despite that exact example being in the card. The developer concluded that prose did not reliably support the comparison and shifted the revision from experience content to control logic.

\paragraph{Explicit ranking and registered checks.}
The revised harness computes a ranking of animals observed in the episode and inserts it into the model request as \texttt{animal\_life\_spans}. A separate check blocks focus on a non-extreme animal when a strictly better candidate is known; it does not intervene for species without a recorded life span. This candidate, denoted R5 in the development record, also included a plant-stage check, stale-record cleanup, and a rule that stops yielding after three repetitions of the same state. Before validation, the developer registered the following life-span checks: the target task v0056 should score 100; each of five other, protected life-span tasks should remain at 100; and the ranking table should be present in the requests for all six episodes, agree with an offline recomputation from raw observations, and place every accepted focus at the extreme requested by the task.

\paragraph{Outcomes and candidate disposition.}
The target and all five protected life-span tasks scored 100. The ranking table was present in all six episodes and matched the offline recomputation. The life-span blocking rule never fired because the model selected the correct answers after receiving the table; the record marked it as not triggered rather than failed. The registered plan also required at least one of the two new blocking rules to fire, with trajectory checks distinguishing non-triggering from failure for the other rule.

The whole R5 configuration was retained as a candidate but was not accepted: neither new blocking rule fired, and two protected tasks from other families regressed. Under the preregistered fallback, the higher-scoring configuration remained a candidate for further validation. R6 evaluated the same configuration without code changes on 52 previously unrun variants under a new preregistered plan and accepted it after all conditions passed. The plant-stage check blocked one incorrect answer. The life-span check remained untriggered because all six life-span tasks were answered correctly; its blocking effect therefore remained unverified. Acceptance applied to the whole configuration, not to independent verification of every rule it contained.